\documentclass[twocolumn]{article}

\usepackage{graphicx} 
\usepackage{array,multirow,graphicx}
\usepackage{float}
\usepackage{tikzsymbols}
\usepackage{hyperref}
\usepackage{authblk}

\title{Challenges and opportunities in portraying emotion in generated sign language}

\author[1]{John C. McDonald}
\author[2]{Rosalee Wolfe}
\author[3]{Fabrizio Nunnari}

\affil[1]{DePaul University, Chicago, IL, USA}
\affil[2]{Institute for Language and Speech Processing, Athena RC, Athens, Greece}
\affil[3]{German Research Center for Artificial Intelligence (DFKI), Saarbr{\"u}cken, Germany}

\date{}

\begin{document}

\maketitle

\begin{abstract}
Non-manual signals in sign languages continue to be a challenge for signing avatars. More specifically, emotional content has been difficult to incorporate because of a lack of a standard method of specifying the avatar's emotional state.  This paper explores the application of an intuitive two-parameter representation for emotive non-manual signals to the Paula signing avatar that shows promise for facilitating the linguistic specification of emotional facial expressions in a more coherent manner than previous methods. Users can apply these parameters to control Paula's emotional expressions through a textual representation called the EASIER notation. The representation can allow avatars to express more nuanced emotional states using two numerical parameters. It also has the potential to enable more consistent specification of emotional non-manual signals in linguistic annotations which drive signing avatars.
\end{abstract}

\section*{Keywords}
Signing avatars, non-manual signals, emotion, sign language animation.

\section{Introduction}
In collaboration with deaf communities, we are developing avatars with the goal of portraying sign languages in a natural, legible manner. If signing avatars are to support the full range of sign language communication, they must be able to portray both linguistic and non-linguistic elements of these rich languages, including the full range of non-manual signals that are communicated by the torso and face of the signer \cite{mcdonald2025considerations}. One key aspect that continues to challenge signing avatars is emotion, especially in the presence of linguistic processes that are portrayed using non-manual facial features. 

From a linguistic standpoint, emotion has traditionally been interpreted as a supralinguistic process, and not included in primary linguistic descriptions or annotations. There are some exceptions to this in recent work \cite{phdthesis} \cite{sharma2024facial}. While such signals are often not included as primary carriers of meaning in linguistic descriptions of signed discourse, they are critical to synthesized sign as they enhance both the meaning and naturalness of the synthesized discourse. Presenting only the lexical and syntactic components of sign strips the discourse of much of the information contained in human communication, and contributes to user impressions of avatar signing as robotic. Written text has a similar problem, which is why much online communication is now done with emotional signals such as ALL CAPS, emoticons and emojis \dSmiley. Luckily, in the case of sign language, we are not dealing with a written form but rather visual portrayal, and so our goal is to layer emotive signals onto the facial motion of the avatar. 

Prior research and feedback have provided valuable insights that inform emotive sign language generation, but also reveal challenges to such a goal. For example, a facial posture can convey different meanings depending on the intent of the signer. Raised eyebrows might indicate the posing of a polar question, but in a different context or in concert, raised eyebrows might indicate surprise \cite{kimmelman2020eyebrow}. Further, grammatical non-manual signals, such as lexical mouth actions \cite{van2009lexically}, can co-occur on the face simultaneously with emotion, but the onset and duration of each process may differ \cite{weast2011american}. 

\begin{figure*}
     \includegraphics[width=\linewidth,keepaspectratio]{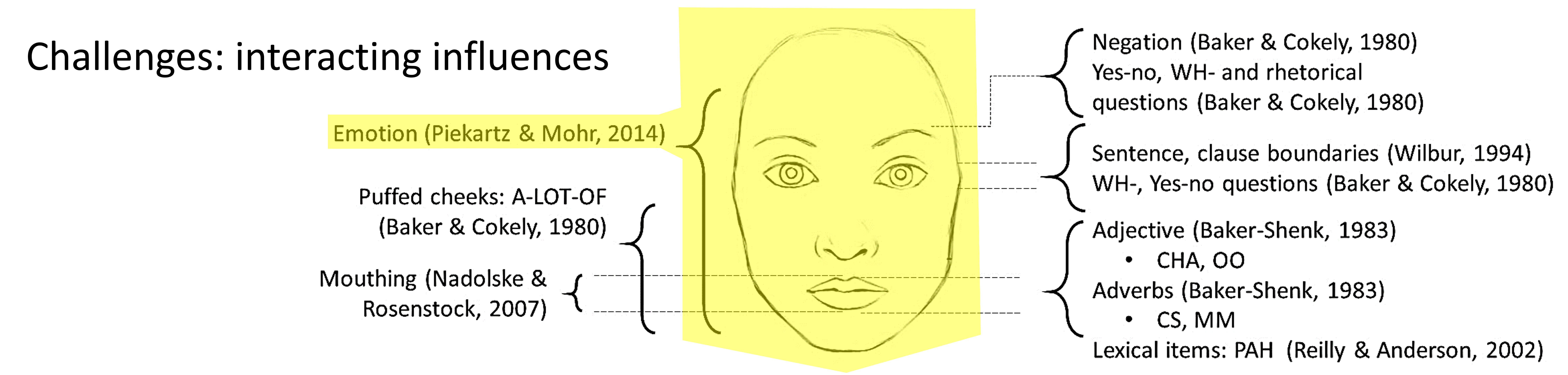}
     \caption{Interacting influences on the signer's face.}
     \label{fig:ManyInfluences}     
\end{figure*}

As shown in Figure~\ref{fig:ManyInfluences}, even at the linguistic level, there may be many asynchronous processes that affect each part of the human face, and these must be synthesized and blended on the face to produce natural discourse. Note that some of the linguistic influences displayed here can, in theory, be isolated in their influence, but all will need to blend with what is occurring on the face emotionally. Further, even those that seem to be isolated, such as the sign PAH in ASL, which means "success" and involves a mandatory mouth movement, can subtly affect the cheeks and even the eyes, see \cite{PAH-Lifeprint}. 

Using a layered approach for organizing the animation preserves the intent and integrity of individual processes, allowing them to combine legibly \cite{schnepp2012combining}. Although this approach has proved capable of portraying several co-occurring processes, there are still open questions, including resolving the competing influences of emotion and mouthing on the lower part of the face, as well as integrating the actions of manual and non-manual channels. Although researchers have made initial attempts to analyze the effect of emotion on the manual channel \cite{hietanen2004perception}, the question of how to incorporate this into sign language generation remains open. This presentation describes ongoing work to represent non-manual emotional signals and identifies challenges to portraying emotion through a signing avatar.

Much of the research in computer science that addresses the recognition and synthesis of facial emotions, relies on Ekman's established list of universal facial expressions (happiness, sadness, fear, surprise, anger, disgust, contempt) \cite{ekman1978facial}. Instead of relying on a discrete model which is limited to seven classes, we experiment with the description and manifestation of emotions expressed in the Pleasure, Arousal, Dominance (PAD) representation described in \cite{bakker2014pleasure}. The PAD model does not use descriptive labels such as sadness or anger, but instead defines emotions via the combination of three values P, A, and D, which are each specified in a closed range. These values control different aspects of facial communication and can be combined to express a wide range of emotion.

\section{Emotion representation and the PAD model}
Modern representation of the facial expression of motion began in 1972 with Paul Ekman's identification of seven basic emotions that all humans have in common, and also identified key elements of facial expressions that communicate these emotions \cite{wyrembelski2014detection}. 
\begin{itemize}
    \item \textbf{Happiness:} Broad smile, eye wrinkles
    \item \textbf{Sadness: } Drooped mouth corners, brow furrow
    \item \textbf{Anger: } Furrowed brow, narrowed eyes, pressed lips
    \item \textbf{Fear: } Wide eyes, raised eyebrows, tense mouth
    \item \textbf{Surprise: } Wide eyes, raised eyebrows, open mouth
    \item \textbf{Disgust: } Wrinkled nose, raised upper lip, brow furrow
    \item \textbf{Contempt: } One raised mouth corner, narrow eyes
\end{itemize}
Importantly, these were originally built as a system for recognizing facial emotion signals rather than for generating them.

Over the last fifty years, many studies have investigated this set of basic emotions and some have called into question the breadth and universality of these expressions, for example  \cite{barrett2019emotional}. Indeed, the original seven were augmented later to include emotions such as embarrassment and pain among others \cite{guerdelli2022macro}, and Ekman himself augmented his original theory with a description of micro-expressions that he found necessary for emotive detection. 

Likewise, while signing avatars have often relied on this representation as the basis for their display of emotion due to its intuitive description and ease of specification, it has become increasingly evident that the original seven emotions are not expressive enough to cover the range and nuance of facial non-manual signals in sign language \cite{sharma2024facial}. Avatars that contain more flexible facial animation systems rely on a set of control bones or control units, such as the MPEG-4 so-called action units in the Facial Action Coding System which was co-designed by Ekman and Friesen \cite{ekman1978facial}. The problem with these systems is that the generation of a specific expression is not nearly as intuitive as the named expressions in Ekman's original or expanded lists. Further, the influence of the Facial Action Coding System (FACS) action units can be difficult to combine given that several may overlap on parts of the face. Finally, it can be difficult to generate new emotions as in some cases the action unit granularity is too coarse to generate some subtleties. 

Some systems address this by introducing enormous numbers of 
controls as in the Metahuman facial rig which strives for ultra-realism which has over 200 controls on 800 joint \cite{Metahuman}. As discussed in \cite{mcdonald2025considerations}, signing avatars need to strike a much more nuanced balance of control and expressivity since they will be driven by linguistic descriptions. So, in this case, we are looking for a flexible but yet constrained model of facial emotion that can be controlled with a few parameters in linguistic annotation.

\begin{figure}
    \includegraphics[width=\linewidth,keepaspectratio]{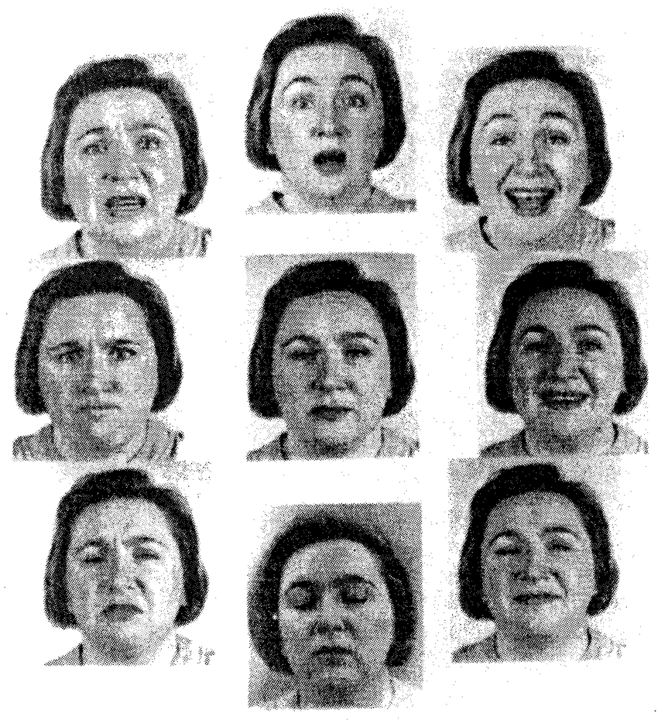}
    \caption{Pleasure (H) and Arousal (A) dimensions (from \cite{russell1997into})}
    \label{fig:PadImages}
\end{figure}

The PAD model, developed by Mehrabian and Russel throughout the 70's and 80's provides a flexible, but highly parsimonious method of representation, based on independent dimensions of emotional impact, to specify facial expressions \cite{mehrabian1980basic}. The three dimensions of the PAD model are 
\begin{itemize}
    \item \textbf{Pleasure/Displeasure:} measures how pleasant an emotion may be. Anger and fear are unpleasant ($P < 0$) whereas joy and excitement are pleasant ($P > 0$).
    \item \textbf{Arousal/Nonarousal:} measures how energized an emotion may be. Among unpleasant emotions, anger has higher intensity or arousal ($A > 0$), while boredom, which is also unpleasant, has a low intensity ($A < 0$). 
    \item \textbf{Dominance/Submissiveness: } measures how controlling/ active vs submissive/reactive the emotion is. Anger would generally expressed in a dominant fashion ($D > 0$) whereas fear is a more reactive/submissive emotion ($D < 0$).
\end{itemize}
The original authors identified two dimensions that are similar to P and A, as the core affect comprising the ``circumplex model'' of emotion experience \cite{russell1980circumplex}. Figure~\ref{fig:PadImages} contains examples of the Pleasure and Arousal of the PAD representation. It is also worth noting that the original Circumplex model only modeled expressions that were on a circle, roughly corresponding to the outer extremes of the P \& A parameters, rather than allowing intermediate expressions throughout the parameter space.

Due to its parsimony and intuitive specifications, the PAD representation has been used in interactive computer games \cite{kleinsmith2012affective}. Further, several projects have applied the PAD model to more general computer-generated avatars. For example, in \cite{lim2007new} these dimensions were applied to 2D animated figures, and in \cite{boukricha2009pleasure} they were applied to 3D animated figures. While the PAD model has been used in a sign language study \cite{reilly1992affective}, its application to signing avatars has received less attention. 

\section{Implementing Pleasure \& Arousal on the Paula Avatar}
This paper proposes the PAD model as a viable candidate for the production of emotion through automatic generation and describes the application of the core affect P \& A dimensions to the Paula avatar. The model supports a more consistent linguistic specification that is suited to text annotation.  In addition, it reports on the results of a recent project (EASIER, \hyperlink{https://www.project-easier.eu}{https://www.project-easier.eu}) that used this representation as the basis for conveying the content of emotions recognized in spoken language to a signing avatar. 

\begin{figure}
     \includegraphics[width=.9\linewidth,keepaspectratio]{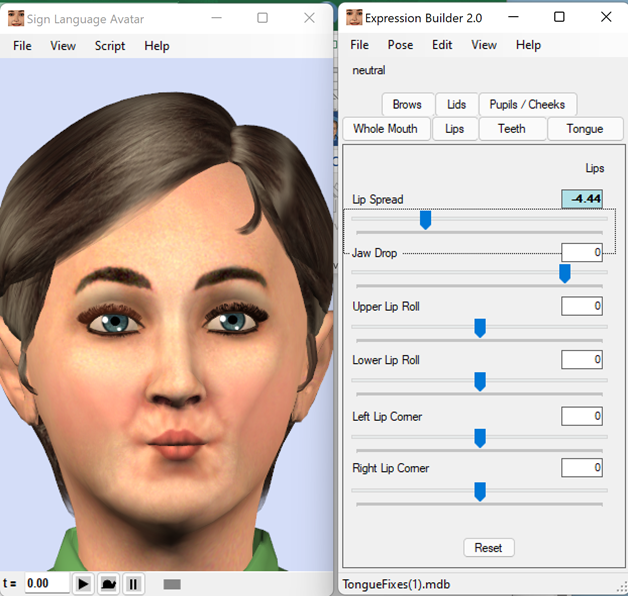}
     \caption{Lip controls for the avatar Paula.}
     \label{fig:PaulaControls}     
\end{figure}

The Paula avatar is equipped with a flexible bone and skin-based facial rig that was originally based on the FACS system cited above~\cite{johnson2022improved}, but which has been expanded beyond the initial action units to encompass facial movements such as fine control of eyelid shape that the original FACS could not express.  An example of the control panels for Paula's face is shown in Figure~\ref{fig:PaulaControls}. For more technical details on the avatar's facial animation system, see \cite{mcdonald2022novel}. In total, there are over 60 controls in this interface, and it gives artists a high degree of range and precision when creating portrayals of emotion.

As discussed in the last section, this is far too fine a control system for linguistic specification and as discussed, the goal is to use this existing facial animation system to implement the core P \& A parameters of the PAD model. Further, to keep potential linguistic use for annotation or analysis as simple as possible, we will restrict the values of each of the two dimensions to +1, -1, and 0. This is actually in keeping with Russel's initial construction of the Circumplex model as the outer extremes of the parameter space.   

\begin{figure}
     \includegraphics[width=1.0\linewidth,keepaspectratio]{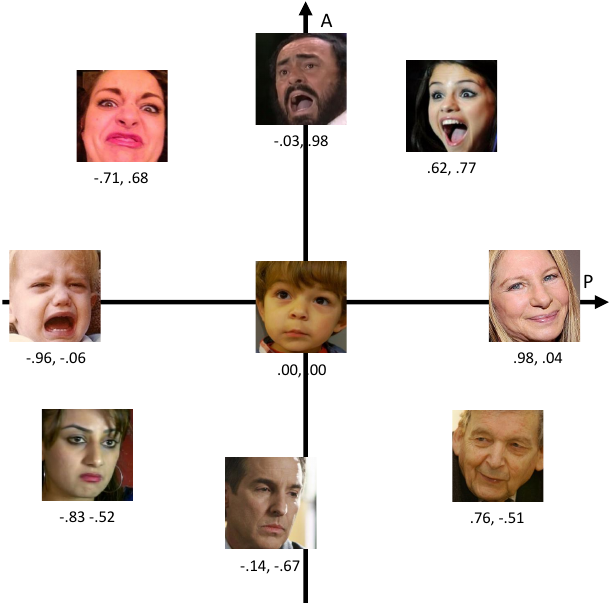}
     \caption{AffectNet pictures that are the closest to the 9 corner cases of the PAD model.}
     \label{fig:AffectNet-references}
\end{figure}

To implement the PA-model, an artist trained in sign language animation used Paula's facial animation interface to create representative expressions for each of the nine extreme poses corresponding to combinations of $\pm1$ or zero in arousal and pleasure (9 \emph{corner cases}).

To give visual references to the artist, we used the AffectNet dataset \cite{mollahosseini_affectnet_2019}: a popular dataset of affective expressions manually annotated for both pleasure and arousal. Figure \ref{fig:AffectNet-references} shows the closest picture to each corner case. For each of the 9 corner cases, we automatically extracted its 10 closest pictures (according to a Euclidean distance). The resulting 90-picture set was provided to the artist as visual inspiration. 

\begin{figure}
     \includegraphics[width=1.05\linewidth,keepaspectratio]{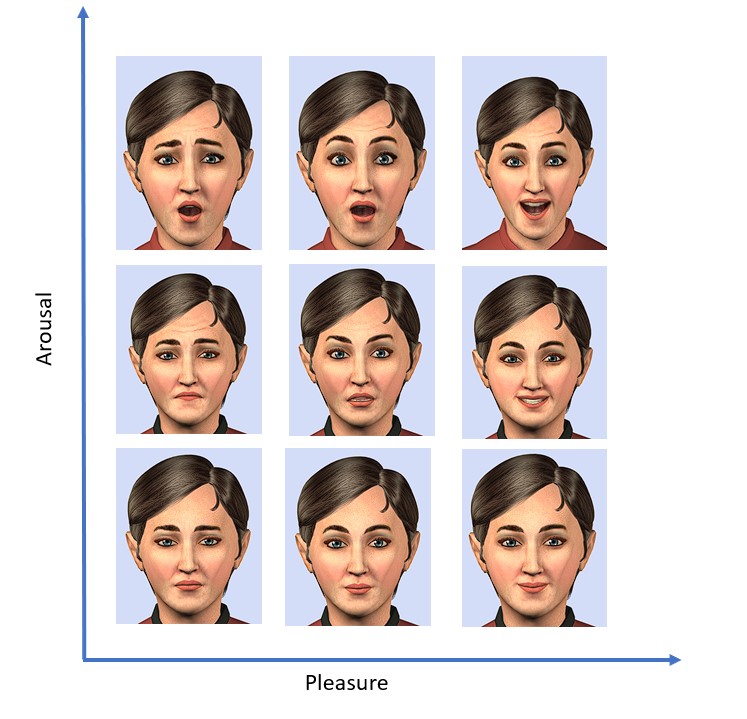}
     \caption{Expressing the P \& A parameters on Paula}
     \label{fig:PadPaula}     
\end{figure}

The resulting Paula expressions are displayed as a three-by-three grid in Figure ~\ref{fig:PadPaula}. Then from a pair of numerical values $(p, a)$, the proper facial expression can be chosen, applied and blended with other facial features as described in \cite{mcdonald2022novel}. The only restriction in the current implementation is that the parameters are limited to integer, i.e. extreme values.

\begin{figure*}
    \includegraphics[width=.92\linewidth,keepaspectratio]{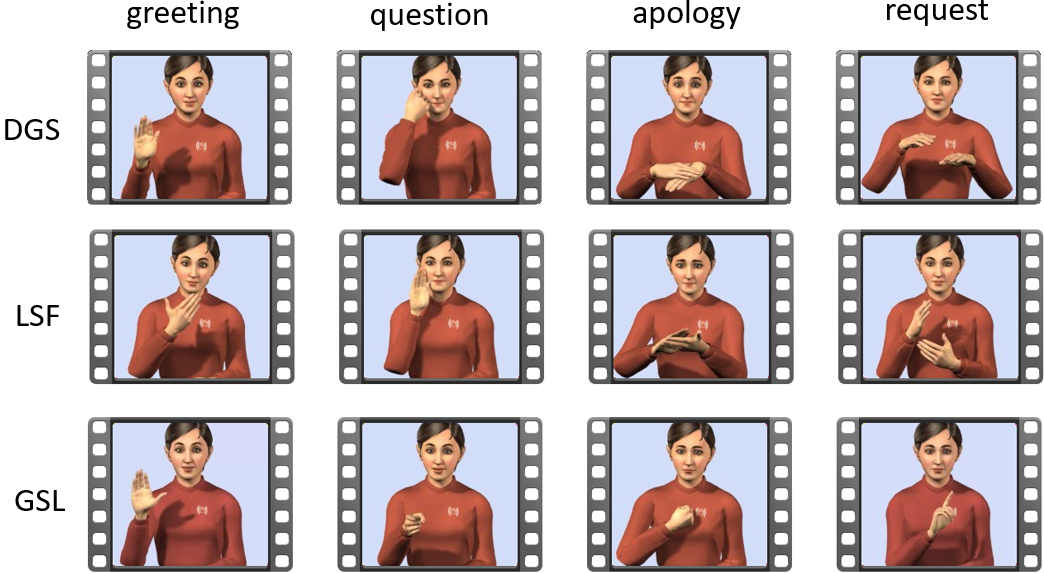}
    \caption{Emotive differences between signed languages}
    \label{fig:CulturalDifferences}     
\end{figure*}

\section{Cultural and Linguistic Differences}
A final challenge that must be addressed by a facial emotive system for signing avatars is the fact that expressions communicating the same basic message with the same basic emotions may be expressed with different facial postures in different signed languages and even in different cultures that share the same signed language.

The EASIER project conducted user evaluations that assessed the quality of the animations, including facial expressions, produced by the Paula avatar.  The test stimuli were five sentences that might be needed in any system that aims to translate between spoken and signed languages. In English, these sentences were as follows.
\begin{enumerate}
    \item Hello, I'm ready to begin.
    \item Could you repeat that?
    \item Sorry, I didn't understand.
    \item Please wait, a response is pending.
    \item Thank you for using our service. Bye!
\end{enumerate}
These sentences were translated by certified translators into Greek (GSL), German (DGS) and French (LSF) sign languages. Aside from the expected differences in manual forms, differences in emotive non-manuals were noted; see Figure~\ref{fig:CulturalDifferences}. Specifically, this is visible in the second column where the same essential question is signed with furrowed brows in DGS and LSF and with raised brows in GSL. Also evident in this figure is that in the third column, the apology facial expression has a very different presentation in the three languages.  Even the intensity of the smile is different among the translations in the first column's ``greeting''. Despite Ekman's observation that there are many commonalities among emotive facial expressions across cultures, there can be significant differences in the subtleties of those expressions. As seen in Figure~\ref{fig:CulturalDifferences},  even the expression used for a specific purpose can differ between cultures. 

In this respect, the PAD system has advantages, as its dimensions are descriptive of the form of the posture rather than a name for an emotion. In the EASIER project, this system was effective for creating facial animation that was deemed appropriate by groups from four different language communities. The method presented here made it quick and easy for annotators to specify non-manual facial signals that the avatar should use in each language.

\section{Conclusion and Future Work}
There are still many open questions for choosing the most effective strategy for mapping PAD values into avatar motion. We currently rely on research from psychology, however, when person-to-person communication relies on sign language, rather than on spoken language, assumptions based on previous research might not hold. It is imperative that next steps include refining avatar portrayal in collaboration with deaf researchers and deaf communities. Most important is to continue the user testing started in the EASIER project both for the legibility of the expressions and also for the ease of use in specifying desired expressions in synthesized discourse.

The current implementation only supports the extreme cases in each dimension. This is not an inherent limitation of the technology, as interpolations of bone rotations may be used to produce intermediate expressions. Rather, this limitation was specifically chosen so that the representation would be easy to specify when creating expressions. In the future, the ability for users to specify intermediate expressions will be expanded. Whether the user should be presented with a continuum of possibilities or rather a discrete selection of choices in each dimension remains to be researched.

In addition, the current model only implements the P \& A dimensions of the PAD model, and this limitation restricts the range of emotion that the avatar can express through these parameters. Adding the ``Dominance'' dimension would require increasing the number of prototype facial poses from 9 to 27, which will require additional animator work.  However, more significant is the fact that among these parameters, P \& A only involve the face in the current theory. The Dominance dimension also involves torso and shoulder actions that complicate the implementation. 

In conclusion, there are many interesting and promising avenues to explore, but only those considered worthy by deaf communities are worth pursuing. Thus, the most important open question is that of collaboration. How can we encourage more deaf researchers and members of the deaf community to participate in avatar research?

\bibliographystyle{plain}
\bibliography{PAD-Annotation,from-nunnari}

\begin{thebibliography}{10}

\bibitem{bakker2014pleasure}
Iris Bakker, Theo Van Der~Voordt, Peter Vink, and Jan De~Boon.
\newblock Pleasure, arousal, dominance: Mehrabian and russell revisited.
\newblock {\em Current psychology}, 33:405--421, 2014.

\bibitem{barrett2019emotional}
Lisa~Feldman Barrett, Ralph Adolphs, Stacy Marsella, Aleix~M Martinez, and
  Seth~D Pollak.
\newblock Emotional expressions reconsidered: Challenges to inferring emotion
  from human facial movements.
\newblock {\em Psychological science in the public interest}, 20(1):1--68,
  2019.

\bibitem{boukricha2009pleasure}
Hana Boukricha, Ipke Wachsmuth, Andrea Hofst{\"a}tter, and Karl Grammer.
\newblock Pleasure-arousal-dominance driven facial expression simulation.
\newblock In {\em 2009 3rd International Conference on Affective Computing and
  Intelligent Interaction and Workshops}, pages 1--7, Amsterdam, 2009. IEEE.

\bibitem{ekman1978facial}
Paul Ekman and Wallace~V Friesen.
\newblock Facial action coding system.
\newblock {\em Environmental Psychology \& Nonverbal Behavior}, 2, 1978.

\bibitem{Metahuman}
Epic Games.
\newblock {MS Windows NT} kernel description.
\newblock \url{https://www.unrealengine.com/en-US/metahuman}, 2021.
\newblock Accessed: 2025-05-27.

\bibitem{guerdelli2022macro}
Hajer Guerdelli, Claudio Ferrari, Walid Barhoumi, Haythem Ghazouani, and
  Stefano Berretti.
\newblock Macro-and micro-expressions facial datasets: A survey.
\newblock {\em Sensors}, 22(4):1524, 2022.

\bibitem{hietanen2004perception}
Jari~K Hietanen, Jukka~M Lepp{\"a}nen, and Ulla Lehtonen.
\newblock Perception of emotions in the hand movement quality of finnish sign
  language.
\newblock {\em Journal of nonverbal behavior}, 28:53--64, 2004.

\bibitem{johnson2022improved}
Ronan Johnson.
\newblock Improved facial realism through an enhanced representation of
  anatomical behavior in sign language avatars.
\newblock In {\em Proceedings of the 7th International Workshop on Sign
  Language Translation and Avatar Technology}, pages 53--58, 2022.

\bibitem{kimmelman2020eyebrow}
Vadim Kimmelman, Alfarabi Imashev, Medet Mukushev, and Anara Sandygulova.
\newblock Eyebrow position in grammatical and emotional expressions in
  kazakh-russian sign language: A quantitative study.
\newblock {\em PloS one}, 15(6):e0233731, 2020.

\bibitem{kleinsmith2012affective}
Andrea Kleinsmith and Nadia Bianchi-Berthouze.
\newblock Affective body expression perception and recognition: A survey.
\newblock {\em IEEE Transactions on Affective Computing}, 4(1):15--33, 2012.

\bibitem{lim2007new}
Mei~Yii Lim and Ruth Aylett.
\newblock A new approach to emotion generation and expression.
\newblock In {\em Proceedings of the Doctoral Consortium, 2nd International
  Conference on Affective Computing and Intelligent Interfaces}, pages
  147--154, Berlin, Germany, 2007. Springer.

\bibitem{mcdonald2025considerations}
John McDonald.
\newblock Considerations on generating facial nonmanual signals on signing
  avatars.
\newblock {\em Universal Access in the Information Society}, 24(1):19--36,
  2025.

\bibitem{mcdonald2022novel}
John McDonald, Ronan Johnson, and Rosalee Wolfe.
\newblock A novel approach to managing lower face complexity in signing
  avatars.
\newblock In {\em Proceedings of the 7th International Workshop on Sign
  Language Translation and Avatar Technology}, pages 67--72, Paris, France,
  2022. European Language Resources Association (ELRA).

\bibitem{mehrabian1980basic}
Albert Mehrabian.
\newblock {\em Basic dimensions for a general psychological theory:
  Implications for personality, social, environmental, and developmental
  studies}.
\newblock Oelgeschlager, Gunn \& Hain, Cambridge, 1980.

\bibitem{mollahosseini_affectnet_2019}
Ali Mollahosseini, Behzad Hasani, and Mohammad~H. Mahoor.
\newblock {AffectNet}: {A} {Database} for {Facial} {Expression}, {Valence}, and
  {Arousal} {Computing} in the {Wild}.
\newblock {\em IEEE Transactions on Affective Computing}, 10(1):18--31, January
  2019.

\bibitem{reilly1992affective}
Judy~S Reilly, Marina~L McIntire, and Howie Seago.
\newblock Affective prosody in american sign language.
\newblock {\em Sign Language Studies}, 75(1):113--128, 1992.

\bibitem{russell1980circumplex}
JA~Russell.
\newblock A circumplex model of affect journal of personality and social
  psychology 39.
\newblock {\em {\'I}6I-I78}, 39(6):1161--1178, 1980.

\bibitem{russell1997into}
James~A Russell.
\newblock Reading emotions from and into faces: Resurrecting a
  dimensional-contextual perspective.
\newblock {\em Psychology of facial expression}, page 295, 1997.

\bibitem{schnepp2012combining}
Jerry~C Schnepp, Rosalee~J Wolfe, John~C McDonald, and Jorge~A Toro.
\newblock Combining emotion and facial nonmanual signals in synthesized
  american sign language.
\newblock In {\em Proceedings of the 14th international ACM SIGACCESS
  conference on Computers and accessibility}, pages 249--250, 2012.

\bibitem{sharma2024facial}
Paritosh Sharma, Camille Challant, and Michael Filhol.
\newblock Facial expressions for sign language synthesis using facshuman and
  azee.
\newblock In {\em 11th Workshop on the Representation and Processing of Sign
  Languages: Evaluation of Sign Language Resources}, pages 354--–360, 2024.

\bibitem{phdthesis}
Robert Smith.
\newblock {\em The role of emotional and facial expression in synthesised sign
  language avatars.}
\newblock PhD thesis, Technological University Dublin, 01 2014.

\bibitem{van2009lexically}
Inge Van~de Sande and Onno Crasborn.
\newblock Lexically bound mouth actions in sign language of the netherlands.
\newblock {\em Linguistics in the Netherlands}, 78:90, 2009.

\bibitem{PAH-Lifeprint}
Dr.~William Vicars.
\newblock Pah: The american sign language (asl) sign for "pah! / finally / at
  last / eureka".
\newblock \url{https://www.lifeprint.com/asl101/pages-signs/p/pah.htm}.
\newblock Accessed: 2025-07-15.

\bibitem{weast2011american}
Traci Weast.
\newblock American sign language tone and intonation: A phonetic analysis of
  eyebrow properties.
\newblock {\em Formational units sign languages}, pages 203--228, 2011.

\bibitem{wyrembelski2014detection}
Adam Wyrembelski.
\newblock Detection of the selected, basic emotion based on face expression
  using kinect.
\newblock {\em
  https://pdfs.semanticscholar.org/6d93/4b68079cbb802fd3bcfac2cae2e5e6d4f7b7.
  (Retrieved 21st Jan. 2018)}, 2014.

\end{thebibliography}

\end{document}